\documentclass{article}

\usepackage{graphicx}        
\usepackage{multicol}        

\usepackage{amsmath,amsfonts, amssymb}

\usepackage[utf8]{inputenc}

\usepackage{todonotes}
\usepackage[scale=0.65]{geometry}

\DeclareMathOperator{\enc}{Enc}
\DeclareMathOperator{\dec}{Dec}

\usepackage{xcolor}
\usepackage{graphicx}
\usepackage{mathrsfs}
\usepackage{float}
\usepackage{multirow}
\usepackage{caption}
\usepackage{subcaption}
\usepackage{tabularx}
\usepackage{authblk}

\title{Reprogramming FairGANs with Variational Auto-Encoders: A New Transfer Learning Model}

\author[1]{Beatrice Nobile\thanks{nobile.1908315@studenti.uniroma1.it}}
\author[2]{Gabriele Santin\thanks{gsantin@fbk.eu}}
\author[1]{Bruno Lepri\thanks{lepri@fbk.eu}}
\author[2]{Pierpaolo Brutti\thanks{pierpaolo.brutti@uniroma1.it}}

\affil[1]{Sapienza University of Rome}
\affil[2]{DIGIS, Bruno Kessler Foundation}

\begin{document}

\maketitle

\begin{abstract}
Fairness-aware GANs (FairGANs) exploit the mechanisms of Generative Adversarial Networks (GANs) to impose fairness on the generated data, freeing them from both disparate impact and disparate treatment. 
Given the model's advantages and performance, we introduce a novel learning framework to transfer a pre-trained FairGAN to other tasks. This reprogramming process has the goal of maintaining the FairGAN's main targets of data utility, classification utility, and data fairness, while widening its applicability and ease of use.
In this paper we present the technical extensions required to adapt the original architecture to this new framework (and in particular the use of Variational Auto-Encoders), and discuss the benefits, trade-offs, and limitations of the new model.
\end{abstract}

\noindent\textbf{Keywords:} FairGAN, Reprogramming, Variational Auto Encoder, Fairness

\section{Introduction}
\label{sec:intro}

The research on Fairness in automated systems has recently advanced considerably, even if recent developments have made the field increasingly complex.
The paper \cite{mehrabi2019survey} lists as many as 23 possible biases in the data, and 6 different types of discrimination that may arise from it, but most importantly they also list 10 different definitions of fairness with no method that can address them all at the same time \cite{pleiss2017fairness}.

Within this complex landscape, FairGANs \cite{xu2018fairgan} offer a simple and actionable mechanism to address biases in the data. Rather than debiasing the data at hand, FairGANs use a GAN-like mechanism to generate new data that must be as similar as possible to the original, but with the additional feature that no evident correlation with the protected attribute is present. As such FairGANs can ensure that the system would be free of both \textit{disparate treatment} and \textit{disparate impact}, i.e., implicit and explicit discrimination in its decision making. 

Since the need for retraining at each usage may be cumbersome at best and unfeasible in many cases where there is not enough data or computing power, this paper aims at creating a transfer learning model for FairGANs. This could lead to a wider adoption of such a method given an increased ease of usage, minimized resource demands, and the ability to re-use knowledge from high-quality data, potentially without loss of accuracy.

To achieve these goals, we make use of Adversarial Reprogramming techniques \cite{elsayed2018adversarial,lee2020reprogramming}), which can retrain the model without the need for the original dataset, and only modifiying a reduced part of the model. 
The biggest issue in applying transfer learning for FairGANs with Adversarial Reprogramming is that the two architectures have been designed to work on different types of data. In fact, while FairGANs have to do with tabular data, Adversarial Reprogramming so far has only been applied to images and sequences \cite{lee2020reprogramming,neekhara2018adversarial}. 
Therefore, the first contribution of this paper is designing a model that is able to perform Reprogramming on tabular data. To do so, we have used Variational Auto-Encoders \cite{doersch2016tutorial} which allow for the necessary dimensional flexibility. The second contribution is understanding whether transfer learning is possible on FairGANs, or if instead direct training is absolutely necessary at each new task. Our findings suggest that balancing all three objectives of FairGANs, data utility and fairness and classification utility, is not trivial when performing transfer learning. Yet, even extreme cases have yielded positive results in most experiments, suggesting that it is indeed possible.

\section{Materials and Methods} 
\label{sec:methods}
We recall the tools used in our model in Section \ref{sec:background}, and introduce the complete solution in Section \ref{sec:model}.



\subsection{Background on GANs, FairGANs, and reprogramming}\label{sec:background}
GANs were first introduced in 2014 \cite{goodfellow2014generative}. Given an initial dataset $\mathcal D:=\{x_i\}$ of certain objects, the idea is to train two neural networks, a generator $G$ and a discriminator $D$, so that $G$ is able to generate, from a random input seed $\bar z$, an object $\bar x$ that is as similar as possible to the true ones, while $D$ is able to classify real and fake objects. The two networks are trained to minimize a minimax loss  \cite{goodfellow2014generative}.

FairGANs \cite{xu2018fairgan} follow the same mechanism but applied now to an extended dataset $\mathcal D:=\{(x_i, y_i, s_i)\}$, where $x_i$ is a certain object, $y_i$ is an associated binary label, and $s_i$ is a binary value representing the membership of $x_i$ in a protected class.
In this case $G$ is seeded with a random input $\bar z$ and a value $\bar s\in\{0,1\}$, and outputs a generated pair $(\bar x, \bar y)$. A first discriminator $D_1$ tries to tell apart the triples $(\bar x, \bar y, \bar s)$ from those found in $\mathcal D$, while a second discriminator $D_2$ tries to classify the pairs $(\bar x, \bar y)$ generated with an input $\bar s=0$ from those generated with an input $\bar s=1$, of course without access to the actual value of $\bar s$.
A modification of the minimax optimization problem forces $G$ to fool both discriminators, thus learning to generate data that are realistic (\emph{data utility}), and for which the protected attribute is hard to identify if not disclosed (\emph{data fairness}). 
Once the training is completed, $G$ can be used to generate a bias-free dataset, that can finally be used to train a classifier that predicts the binary label $y$ for a previously unseen object $x$, without relying on $s$. 



In order to perform transfer learning with a pre-trained FairGAN, we additionally use Adversarial Reprogramming techniques \cite{elsayed2018adversarial}. 
In general terms, given a model $f:X\to Y$ between sets $X, Y$ and a target model $g:X'\to Y'$ between sets $X', Y'$, the goal is to modify the input and output of $f$ in order to mimic the response of $g$. A typical example is the modification of a pre-trained classifier that maps a set of images $X$ to suitable labels $Y$, in order to obtain a new classifier $g$ that works on a different set of images $X'$ of possible different resolution, with other labels $Y'$. This goal is achieved by training two parametric functions $h_f:X'\to X$, $h_g:Y\to Y'$ so that the input-output modified model $h_g\circ f\circ h_f:X'\to Y'$ approximates $g$.
This general technique has been applied to GANs in \cite{lee2020reprogramming}, where an existing generator-discriminator pair $(G, D)$ is updated to generate objects of the same original dimension, but following a different distribution. An input modification function $h_f$ is applied to obtain a new generator $G':= G\circ h_f:Z'\to X$, where $Z'$ is a second set of input noise seeds. Since the GAN training step requires the join optimization of the generator and discriminator, the new discriminator $D'$ is instead trained from scratch, 
therefore moving away from an adversarial attack paradigm towards one of transfer learning.



\subsection{Reprogramming FairGANs}\label{sec:model}

We assume to have a dataset $\mathcal D:=\{(x, y, s)\}\subset X\times Y\times \{0,1\}$, where $X$ are tabular data, and initially train a classifer $\overline C:X\to Y$, that will remain fixed. Additionally, we train a standard VAE $\overline\enc, \overline\dec$ on $X \times S$, i.e., $\overline\dec(\overline\enc((x, s))) := \left(G_x(x, s), G_s(x,s)\right)\approx (x, s)$ for all $(x,s)\in X\times S$. This will serve as the base for the construction of the generator and allow us to tackle reprogramming with dimension expansion, which is a notoriously hard to solve problem \cite{lee2020reprogramming}.

We then consider a second target dataset $\mathcal D':=\{(x', y', s')\}\subset X'\times Y\times \{0,1\}$ with tabular input $X'$. We assume that the columns of $X'$ are either a subset or a superset of those of $X$. Observe that a special setting of interest, which we will explore in the following, is the case when $X=X'$ or $Y=Y'$.
Our goal is to adapt the classifier $\overline C$ to $\mathcal D'$ under two constraints: creating realistic data, and creating bias-free data.

To reprogram with the realism constraint only, we define a generator $G:(X',S')\to (X,S)$ defined as $G:=\overline\dec \circ \enc$ which maps the new columns to the old ones. We also define a discriminator $D:X\to\{0,1\}$. Observe that the decoder part of the generator is kept fixed. Instead, we train $\enc$ with two goals: we want to have $\overline C(G_x(x'))\approx y'$, i.e., the classification accuracy is preserved, and we want that  $D_1$ can't distinguish $G(X'\times S')$ from $X'\times S'$. Observe that $D_1$ acts only on the columns of $G(X')$ that are in common with those of $X'$, so that it can be trained by having access to $X'$ only, and not to the original set $X$.
That is, the discriminator should discriminate only those columns that represent the current data.
This goal is realized by the optimization of $\enc$ and $D_1$ by minimization of a loss composed by the usual GAN loss with $\ell_2$ norm penalty, plus the cross entropy loss for classification. The balance between the two losses is controlled by a parameter $\gamma\geq 0$ (large $\gamma$ means large importance to classification).
In this way we realize a transfer of the classifier from $\mathcal D$ to $\mathcal D'$, in a way that preserves a realism condition on the generated data. Observe that even without the realism constraint, this \emph{tabular reprogramming} is an unexplored field for adversarial reprogramming.

When adding a fairness constraint, we obtain the full \emph{FairGAN reprogramming} model, which differs solely for the presence of additional discriminators. 
In this case, a second discriminator $D_2:G(X'\times S')\to S'$ tries to distinguish between the generated columns that are generated with an original input with $s'=0$ or $s'=1$. This second discriminator is trained together with $\enc$ and $D_1$, by adding to the loss a second GAN-discriminator loss term controlled by a second parameter $\delta\geq 0$, where $\delta=0$ means that $D_2$ is deactivated.
With this design, the reprogrammed FairGAN guarantees data fairness because the ''fairness'' discriminator $D_2$ forces the generator to produce data that is independent of the sensitive information.  

\section{Results}
\label{sec:results}
For our experiments we use two datasets, COMPAS Recidivism Racial Bias \cite{Angwin2016} comprising after cleaning 60843 observations and 15 columns, and a Candidate Selection dataset \cite{ispass:2020} comprising 31349 observations and 10 columns. 
We consider as target labels $y$ either \textit{gender} (both datasets), or \textit{ethnic code} (Caucasian or not, in COMPAS) and \textit{selected} (hired or not - Candidate).

Table \ref{tab:summary} reports the accuracies of these four classifiers $\overline C$, that we consider as baselines for our experiments. For the reprogramming tasks we consider four scenarios: i) We don't change neither dataset nor task (Same Dataset Same Task, SDST); ii) We change the task but not the dataset (Same Dataset Different Task, SDDT); iii) We change dataset while maintaining the same task (Different Dataset Same Task, DDST); iv) We change both dataset and task (Different Dataset Different Task, DDDT). 
We remark that passing from COMPAS to Candidate requires a dimensionality reduction, while the vice versa requries a dimensionality expansion.

In all experiments, we use the architecture of Section \ref{sec:model}.
The latent dimension is set to $10$, which is the smallest value that does not affect the classification accuracy in the results of Section~\ref{sec:tabresults}.

\begin{table}
\centering
\begin{footnotesize}
\begin{tabular}{ccccc}
\hline
&COMPAS-Gender & COMPAS-Ethnic code & Candidate-Gender&Candidate-Selected\\
\hline
Accuracy & 82.8 & 67.2 & 78.8 & 71.7\\
\hline
\end{tabular}
\end{footnotesize}
\caption{Accuracy of the classifiers trained on different dataset and target labels. 
}\label{tab:summary}
\end{table}

\subsection{Tabular Reprogramming}\label{sec:tabresults}
We first test simply tabular reprogramming, i.e., without fairness constraint. In other terms, the only objective we are trying to achieve is to efficiently fool the classifier in order to maintain high accuracy, regardless of how we generate the data.
We consider the classifier trained on the Candidate dataset to predict the label \emph{gender}, and report the results of the four transfer learning scenarios SDST, SDDT, DDST, DDDT in the left part of Table~\ref{tab:results}. The model works very well in three out of four cases, even outperforming the baseline. Observe that this is possible since the architecture includes now also the optimizable encoder, and thus the overall model has more degrees of freedom. Only in the DDDT setting, which is ideed the more challenging, we observe a limited reduction of  $4\%$ in the accuracy. Thus, we can affirm that we have successfully reprogrammed across tabular datasets.



\begin{table}
\centering
\begin{footnotesize}
\begin{tabular}{l c c c c| c c c c c}
& SDST & SDDT & DDST & DDDT  & $\gamma$ &SDST & SDDT & DDST & DDDT \\
\hline
Reprogramming & 78.9 & 78.7 & 79.1 & 74.5 &  0.5 & 74.2 \textcolor{green}{\checkmark} & 57.6 \textcolor{green}{\checkmark} & 61.1 \textcolor{green}{\checkmark} & 52.7 \textcolor{green}{\checkmark} \\
&&&&&100 & 76.3 \textcolor{red}{$\times$} & 64.8 \textcolor{red}{$\times$} & 69.1 \textcolor{green}{\checkmark} & 64.8 \textcolor{green}{\checkmark} \\ 
Baseline & 78.8 & 78.8 &78.8 & 78.8 & & 82.8\phantom{$\times$} & 67.2\phantom{$\times$} &78.8 \phantom{$\times$}& 71.7\phantom{$\times$} \\
\end{tabular}
\end{footnotesize}
\caption{Accuracy results of tabular reprogramming (left, see Section~\ref{sec:tabresults}) and GAN reprogramming (right, see Section~\ref{sec:ganresults}), compared to the corresponding baseline models obtained by direct training.}\label{tab:results}
\end{table}

\subsection{GAN and FairGAN Reprogramming}\label{sec:ganresults}

We now add the data realism constraint, so that we both reprogram the classifier and require the generated data to be similar to the input one. As described in Section \ref{sec:model}, a parameter $\gamma$ balances the importance of classification and of data realism, where larger values of $\gamma$ increase the relative importance of classification.
The setting of $\gamma$ is itself a challenging task. Cross validation over the values of $\gamma$ in combination with the learning rate $\eta$ have identified a values $\gamma=0.5$, $\eta=10^{-3}$. However, preliminary investigations have shown that reasonable results may be achieved with $\gamma\in [10^{-1}, 10^2]$. We report the results in Table~\ref{tab:results}, where we also report the case of $\gamma=0.5$ for comparison. 
Each experiment reports also a green check or a red cross to indicate if the data realism constraint is met. In these preliminary experiments, this is checked by comparing the histograms of the distributions of the single true and generated columns.
Unsurprisingly, moving across datasets seems to be more challenging for our model, but we manage to keep a relatively small decay in the accuracy, especially considering the semantic distance between the two datasets and the respective tasks.
Adding the fairness constraint we can finally test the entire FairGAN model, where we tested the model on several randomly selected columns used as the protected attribute. 
In this case the loss comprises the three terms of classification accuracy, realism, and fairness, and their relative importance is controlled now by the two hyperparameters $\delta$ and $\gamma$. To measure the success of the fairness constraint, we check if the validation accuracy of the discriminator is close to 50\% on a validation set, meaning that the protected attribute is not predictable from the cleaned data.
In this case, no value of the two parameter give a sufficiently accurate result in SDST and SDDT, while for DDST and DDDT we managed to obtain a validation accuracy close to $50\%$. This apparently counter-intuitive result may be due to preliminary nature of our experiments, and will be addressed in future research. 

\section{Conclusions}
In this work we introduced for the first time a framework for the reprogramming FairGANs on tabular data, i.e., the possibility of using a pre-trained FairGAN and apply it to other datasets and/or tasks, maintaining acceptable levels of \textit{data utility}, \textit{classification utility} and \textit{data fairness}. 
Our novel architecture leverage VAEs as a basic building block, and combines it with two dicriminators which are trained with a GAN loss to forse the generation of data that meets the desired constraints.
Preliminary experiments on two benchmark datasets have shown that the model can be effectively trained to reprogram FairGANs on tabular dataset. Nevertheless, the incremental addition of multiple constraints makes the training increasingly challenging. Even if the experiments lean towards positive results, some issues remain unanswered, especially in the design of an effective balancing mechanism between the various components of the model.

\bibliographystyle{abbrv}
\bibliography{mybibliography.bib}

\end{document}